\documentclass{article}
\usepackage{microtype}
\usepackage{graphicx}
\usepackage{subcaption}
\usepackage{booktabs}
\usepackage{hyperref}
\usepackage{algorithmic}
\usepackage[accepted]{icml2024}
\usepackage{amsfonts}
\usepackage{amsopn}
\usepackage{epstopdf}
\usepackage{graphicx}
\usepackage{amsmath}
\usepackage{amssymb}
\usepackage{mathtools}
\usepackage{amsthm}
\usepackage[capitalize,noabbrev]{cleveref}
\theoremstyle{plain}

\theoremstyle{definition}

\theoremstyle{remark}

\usepackage[textsize=tiny]{todonotes}
\usepackage{bm}
\usepackage{xspace}
\usepackage{multicol}
\usepackage{breqn}
\usepackage{placeins}
\input{helpers}
\icmltitlerunning{Black-box Approximation and Optimization}
\begin{document}
\pdfstringdefDisableCommands{%
  \def\\{}%
  \def\texttt#1{<#1>}%
}
\twocolumn[
\icmltitle{Black-Box Approximation and Optimization\\with Hierarchical Tucker Decomposition}

\begin{icmlauthorlist}
\icmlauthor{Gleb Ryzhakov}{sk}
\icmlauthor{Andrei Chertkov}{airi,sk}
\icmlauthor{Artem Basharin}{airi,sk}
\icmlauthor{Ivan Oseledets}{airi,sk}
\end{icmlauthorlist}

\icmlaffiliation{sk}{Skolkovo Institute of Science and Technology, Moscow, Russia}
\icmlaffiliation{airi}{Artificial Intelligence Research Institute (AIRI), Moscow, Russia}

\icmlcorrespondingauthor{Gleb Ryzhakov}{g.ryzhakov@skoltech.ru}

\icmlkeywords{Black-box optimization,%
    black-box approximation,%
    gradient-free method,%
    low rank representation,%
    hierarchical Tucker decomposition}

\vskip 0.3in
]

\printAffiliationsAndNotice{} 

\begin{abstract}
We develop a new method HTBB for the multidimensional black-box approximation and gradient-free optimization, which is based on the low-rank hierarchical Tucker decomposition with the use of the MaxVol indices selection procedure. Numerical experiments for 14 complex model problems demonstrate the robustness of the proposed method for dimensions up to 1000, while it shows significantly more accurate results than classical gradient-free optimization methods, as well as approximation and optimization methods based on the popular tensor train decomposition, which represents a simpler case of a tensor network.
\end{abstract}

\setcounter{footnote}{2} 
\section{Introduction}
    \label{s:intro}
    Many physical and engineering models can be represented as a real function (output), which depends on a multidimensional argument (input) and looks like
\begin{equation}\label{eq:function}
y = \ff(\vx) \in \set{R},
\quad
\vx = [x_1,\, x_2,\, \ldots,\, x_d]^T \in \Omega \subset \set{R}^d.
\end{equation}
Such functions often have the form of a black-box (BB), \ie, the internal structure and smoothness properties of $\ff$ remain unknown.
Its discretization on some multi-dimensional grid results in a multidimensional array (tensor\footnote{
    By tensors we mean multidimensional arrays with a number of dimensions $d$ ($d \geq 1$).
    A two-way tensor ($d = 2$) is a matrix, and when $d = 1$ it is a vector.
    For scalars we use normal font, we denote vectors with bold letters and we use upper case calligraphic letters ($\tens{A}, \tens{B}, \tens{C}, \ldots$) for tensors with $d > 2$.
    The $(n_1, n_2, \ldots, n_d)$th entry of a $d$-way tensor $\tens{Y} \in \set{R}^{N_1 \times N_2 \times \ldots \times N_d}$ is denoted by $\tens{Y}[n_1, n_2, \ldots, n_d]$, where $n_k = 1, 2, \ldots, N_k$ ($k = 1, 2, \ldots, d$) and $N_k$ is a size of the $k$-th mode. Mode-$k$ slice of such tensor is denoted by $\tens{Y}[n_1, \ldots, n_{k-1}, :, n_{k+1}, \ldots, n_d]\in\mathbb R^{N_k}$.
}) $\ty \in \set{R}^{N_1 \times N_2 \times \ldots \times N_d}$ that collects all possible discrete values of the function \eqref{eq:function} inside the domain $\Omega$, i.e.,
\begin{equation}\label{eq:tensor-function}
\ty[n_1,\, n_2,\, \ldots,\, n_d]
=
\ff\bigl(
    x_1^{(n_1)}\!, \, x_2^{(n_2)}\!, \, \ldots, \, x_d^{(n_d)}
    \bigr).
\end{equation}
Storing such a tensor often requires too much computational effort, and for large values of the dimension $d$, this is completely impossible due to the so-called curse of dimensionality (the memory for storing data and the complexity of working with it grows exponentially in $d$).
To overcome it, various compression formats for multidimensional tensors are proposed: Canonical Polyadic decomposition aka CANDECOMP/PARAFAC (CPD)~\cite{harshman1970foundations}, Tucker decomposition~\cite{tucker1966some}, Tensor Train (TT) decomposition~\cite{oseledets2011tensor}, Hierarchical Tucker (HT) decomposition~\cite{hackbusch2009new,ballani2013black}, and their various modifications.
These approaches make it possible to approximately represent the tensor in a compact low-rank (i.e., low-parameter) format and then operate with the compressed tensor.

The TT-decomposition is one of the most common compression formats~\cite{cichocki2016tensor, cichocki2017tensor}.
There is an algebra for tensors in
the TT-format (i.e., TT-tensors):
we can directly add and multiply TT tensors,
truncate TT tensors (reduce the so-called TT-rank, i.\,e, the number of storage parameters), integrate and contract TT tensors.
It is important that effective algorithms have been developed~\cite{kapushev2020tensor, ahmadiasl2021cross, chertkov2023black} for approximating BB like~\eqref{eq:function} and~\eqref{eq:tensor-function} in the TT-format, that is, for constructing an approximation (surrogate model) using only a small number of explicitly computed BB values.
In recent years, new efficient algorithms have also been proposed~\cite{sozykin2022ttopt, nikitin2022quantum, chertkov2023tensor} for the second important problem associated with gradient-free optimization of such BB, that is, finding an approximate minimum or maximum value based only on queries to the BB.
Such TT-based methods of surrogate modeling (in particular, the TT-cross algorithm~\cite{oseledets2010ttcross}) and gradient-free optimization (in particular, the TTOpt algorithm~\cite{sozykin2022ttopt}) have shown their effectiveness for various multidimensional problems, including compression and acceleration of neural networks, data processing, modeling of physical systems, etc.

However, the TT-decomposition is one of the simplest special cases of a tensor network: it is a linear network or a degenerate tree, and it has a number of limitations related to weak expressiveness and instability for the case of significantly large dimensions.
The HT-format is potentially more expressive and robust~\cite{buczyska2015hackbusch}; thus, it makes it possible to approximate more complex functions with fewer parameters.
Taking into account TT-Cross and TTOpt algorithms which use the well-known MaxVol approach~\cite{goreinov2010how, mikhalev2018rectangular}, in this work we develop new methods of surrogate modeling and gradient-free optimization based on the HT-format,
and our main contributions are the following:
\begin{itemize}
    \item we develop a new black-box approximation method HT-cross based on the HT-decomposition and the rectangular MaxVol index selection procedure;
    
    \item we develop a new gradient-free optimization method HTOpt based on the HT-decomposition and the rectangular MaxVol index selection procedure;

    \item we implement the proposed HT-cross and HTOpt algorithm as a unified method HTBB for surrogate modeling and optimization of multidimensional functions given in the form of a black-box and share it as a publicly available python package;\footnote{
        The program code with the proposed approach and numerical examples, given in this work, is publicly available in the repository
        \url{https://github.com/G-Ryzhakov/htbb}.
    }
    
    \item we apply our approach HTBB to $14$ different complex model functions with input dimensions up to $1000$ and demonstrate its significant advantage in the accuracy and robustness for the same budget in comparison with the TT-cross method for approximation problems and with the TTOpt, and well-known classical gradient-free SPSA and PSO methods for optimization problems.
\end{itemize}

\section{Hierarchical Tucker Decomposition}
    \label{sec:ht}
    By Hierarchical Tucker (HT), we mean a tensor tree that is not necessarily balanced~\cite{ballani2013black}.
Let us describe this concept in detail in the context of our work. 
HT is such a low-parameter decomposition of a $d$-way tensor,  which is a hierarchical contraction of $3$-way tensors and $2$-way tensors, ordered in the form of a binary tree.

Consider a binary tree~--- a graph without cycles, 
where every \defin{node} (except the root one) has a \defin{parent} and at most two \defin{children}.
In what follows, we consider trees where each node has either 2 or 0 children.
We call a node without children a \defin{leaf}.
We denote the depth of the tree by~$L$, and the number of nodes at level~$l$ (starting from the root node) by $\lambda_l$; note that for a balanced tree, $\lambda_l=2^{l-1}$ is satisfied.
With each node, we associate a \defin{core} tensor, i.e., a 2-way tensor with the leaves, 
and 3-way tensors with all others 
(for the root core we add a dummy dimension of the length~$1$).

The number of leaves~$d$ determines the dimensionality of the considered tensor $\mathcal Y$, 
which is represented in the described tree structure, i.e., $\tens Y\in\dims{N_1 \times N_2\times\cdots\times N_d}$, where $N_j$ is the size of $j$th mode.
The dimensions of the cores are as follows. 
Leaves dimensions correspond to the dimensionality of the tensor $\tens Y$:
each core~$\G Lj$ that is associated
with a leaf node with number~$j$ satisfies $\G Lj\in\dims{r^{(L)}_j\times N_j}$.
The dimensions of the non-leaves cores match the dimensions of their children:
if core $\G lj\in\dims{r^{(l+1)}_{j_1}\times r^{(l)}_{j}\times r^{(l+1)}_{j_2}}$
for $1\leq l< L$, then its child $\G {l+1}{j_1}$ and $\G {l+1}{j_2}$ have such dimensions that
$\G {l+1}{j_1}\in\dims{r_1\times r^{(l+1)}_{j_1}\times r_2}$ 
and
$\G {l+1}{j_2}\in\dims{r_3\times r^{(l+1)}_{j_2}\times r_4}$.
The numbers $r^{(i)}_j$
are called \defin{ranks} of the HT decomposition.
For the core of the root node it is hold $\G 11\in\dims{r^{(2)}_{j_1}\times1\times r^{(2)}_{j_2}}$.
In the case of notations related to tree nodes,
the index at the top in parentheses denotes the level of the tree~$l$, it varies from~$1$ to~$L$ ($L=\ln d$ for the balanced tree),
and with the index at the bottom we denote the numbering within this level of the tree, this numbering is not fixed and is arbitrary.

To calculate the value of the tensor~$\tens Y$ represented in the HT-format at a given index~$I$, we perform the following iterative procedure.
We associate a vector~$\nb lj$ with each node, which is recursively defined as
\begin{equation*}
\label{eq:calcval_1}
\nb lj=
\sum_{i=1}^{r_1}
\sum_{k=1}^{r_2}
\G lj[i,\,:,\,k]\cdot
\nb {l+1}{j_1}[i]\cdot
\nb {l+1}{j_2}[k],
\end{equation*}
where the vectors 
$\nb {l+1}{j_1}$ and $\nb {l+1}{j_2}$ are vectors associated with children of the current node;
$r_1=\nr{l+1}{j_1}$ and
$r_2=\nr{l+1}{j_2}$ are the corresponding ranks.
For a leaf node,
its corresponding vector~$\nb Lj$ depends on the given index~$I$ as 
$ 
\label{eq:calcval_2}
\nb Lj=
\G Lj\bigl[:,\,I[j]\bigr].
$ 
Finally,
the resulting tensor value at index~$I$ is equal
to the value of the single element of the vector~$\nb11$
associated with the root node, i.e.,
$ 
    \tens Y[I]=\nb11[1].
$ 
Note that this procedure is easily parallelized naturally since vectors~$b$ of the same level in different parts of the tree are calculated independently.
Moreover, the HT-format is more expressive and robust~\cite{buczyska2015hackbusch} than simpler forms of tensor networks (for example, the well-known TT-decomposition), which makes it potentially possible to build the low-rank approximation for complex functions.
\section{Proposed Approach}
    \label{sec:scheme}
    Due to the potential strengths of the low-rank HT-decomposition for high-dimensional applications described in the previous section, it seems important to develop new approximation and optimization methods based on it.
We are inspired by a simpler, but carefully designed TT-format and implement analogues of the known methods TT-cross and TTOpt on its basis for the HT-decomposition.
The TT-cross algorithm~\cite{oseledets2010ttcross} adaptively calls the BB and iteratively builds the TT-surrogate until a given accuracy is reached or the BB access budget is exhausted.
During this construction, the so-called Maximum Volume submatrix search (MaxVol) procedure~\cite{goreinov2010how} is used to find a close to the dominant matrix of the tensor unfolding.
Thus, this matrix contains values close to the maximum modulus values of the tensor.
This effect can be used to find the quasi-maximal element in the tensor, and the corresponding algorithm is called TTOpt~\cite{sozykin2022ttopt}.
Further in this section, we successively describe our algorithms for approximation (HT-cross) and optimization (HTOpt) in the HT-format.

\subsection{Upper and Down Indices}

The key concept that is used for both the approximation and optimization algorithm is to associate index sets with each link between nodes.
Each link between node~$\nN{l-1}{m}$ and its child~$\nN lj$ have \defin{down}~$\nid lj$ and \defin{upper}~$\niu lj$ indices and corresponding values~$\nvd lj$ and~$\nvu lj$ of this indices.
Since each link is unambiguously defined by the child node~$\nN lj$ it is part of, the index notations are similar to this children node notation and sometimes we refer to these indices as being associated with the child node rather than a relation.

\begin{figure*}[htb]
    \centering
    \includegraphics[width=0.95\linewidth]{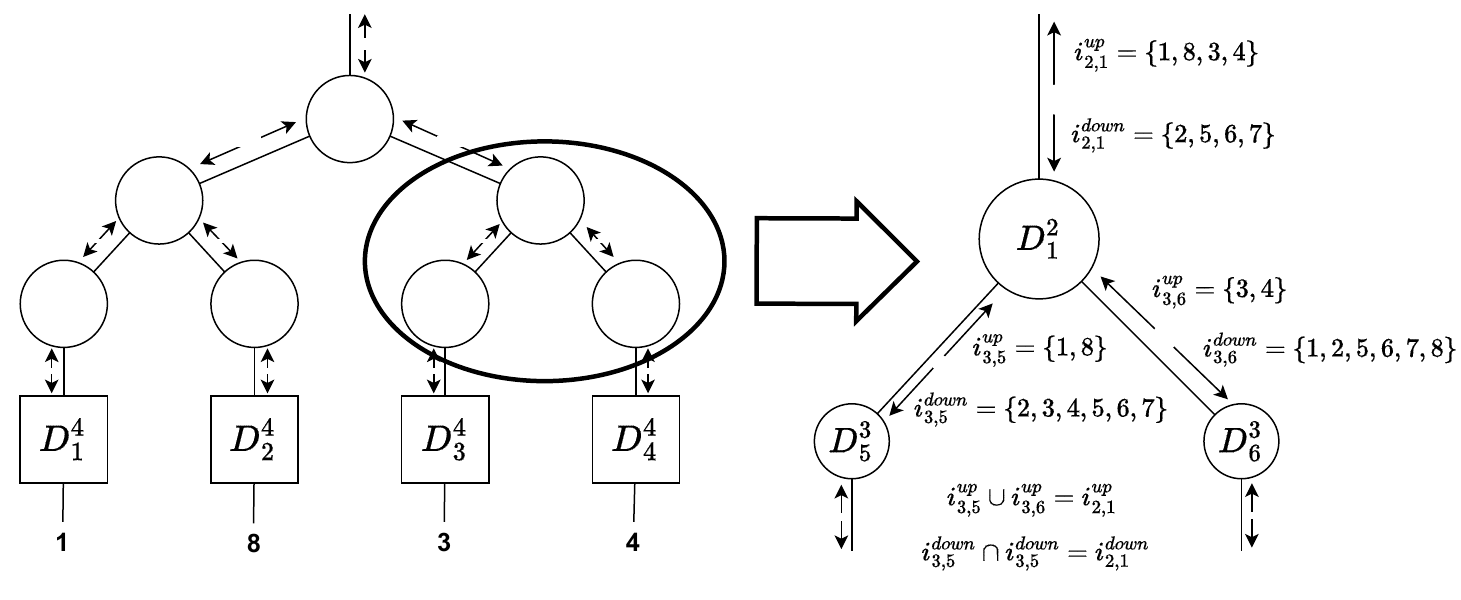}
    \caption{Examples of upper and down indices and their values
        for $\tens Y\in\dims {
            N_1\times
            N_2\times
            N_3\times
            N_4\times
            N_5\times
            N_6\times
            N_7\times
            N_8
        }$ with
        $N_1=N_2=N_5=N_6=N_7=2$, 
        $N_3=N_4=3$, 
        and $N_8=10$.
    }
    \label{fig:indecis}
\end{figure*}

Down~$\nid lj$ and upper~$\niu lj$ indices depend only on their position and are fixed during initial tree construction according to the following recursive rule.
Each leaf node~$\nN Lj$ has upper index~$\niu Lj=\{j\}$ containing one element equal to the element number of the tensor index element that is associated with this leaf node.
Each non-leaf node~$\nN lj$ except the root one has an upper index consisting of the union of the elements of the upper indices of all its children: $\niu lj=\niu {l+1}{j_1}\cup\niu {l+1}{j_2}$.
For all the cases, down indices are equal to the set difference between all tensor indices and upper indices: $\nid lj = \{\fromoneto d\} \setminus \niu lj$.
Since the root node is not a child, we do not associate indexes with it.
The down indices of the left child of the root node and their values are equal to the upper indices of the right child and their values, respectively, and vice versa.
Please see \cref{fig:indecis} for relevant illustration.

\begin{figure*}[htb!]
    \centering
     \begin{subfigure}[b]{0.48\textwidth}
         \centering
         \includegraphics[width=0.99\textwidth]{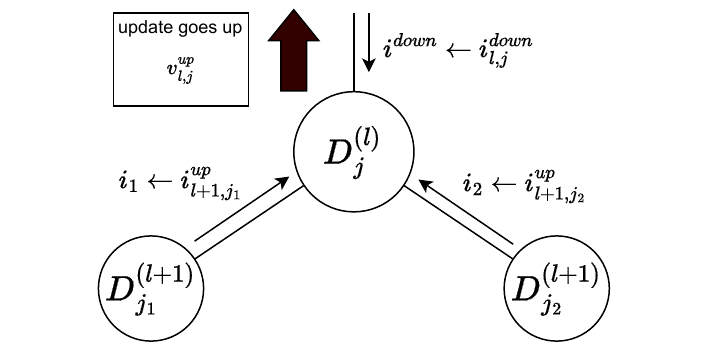}
         \caption{Update upper indices values.}
         \label{fig:update idx up}
     \end{subfigure}
     \hfill
     \begin{subfigure}[b]{0.48\textwidth}
         \centering
         \includegraphics[width=0.99\textwidth]{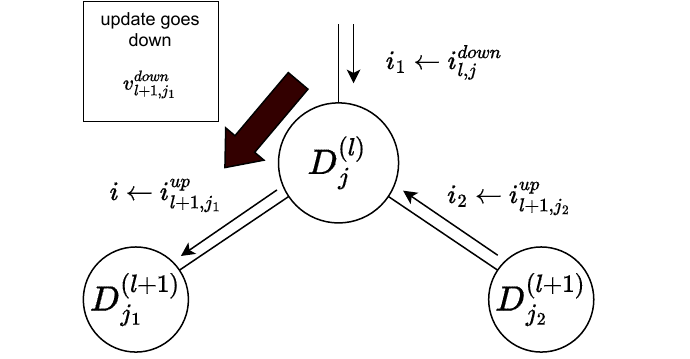}
         \caption{Update down indices values.}
         \label{fig:update idx down}
     \end{subfigure}
    \caption{Algorithm~\ref{alg:update idx} inputs for the cases of upper and down indices values update. On the left: when updating upwards, the indices forming the rows of~$A$ are calculated based on the upper indices on the links below ($i_1$ and $i_2$) and their values, and the indices~$i$ (and their values~$v$) forming the row of the matrix~$A$ consists of the down indices of the link above and their values. The values of the upper indices associated with the link above are updated. On the right: similar updating but with slight changes occurs when moving downwards.}
    \label{fig:update idx}
\end{figure*}

\begin{figure}[ht]
    \centering
    \includegraphics[width=\linewidth]{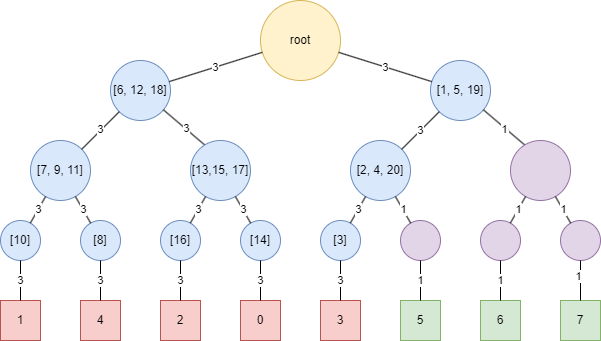}
    \caption{Examples of a path for the traversal procedure.
    The task is 5-dimensional, so indices 5, 6, and 7 (green boxes) 
as well as their parents (purple circles) are never visited.
Ranks of all links,
except for those leading to inactive indices,
are equal to 3.
}
    \label{fig:path}
\end{figure}

The values of the upper~$\nvu lj$ and down~$\nvd lj$ indices change dynamically and the manner and sequence of their change is the subject of this study.
These values~$\nvu lj$ and~$\nvd lj$  represent sets of size equal to the rank, associated with the corresponding node:
$
\left| \nvu lj \right|
=
\left| \nvd lj \right|
=
\nr lj
$.
Each element of this set is a vector with values of indices stored in the corresponding ($\niu lj$ or $\nid lj$) index set.
The main goal of the iterative search for index values (the detailed implementation of which will be described below) is to find the submatrix of maximum volume at the intersection of the given indices.
Finding a submatrix of maximal volume serves two purposes: first, we can more accurately reconstruct the original matrix using it, and second, we expect that this matrix has elements close to maximal in modulo.
Let us elaborate on the construction of this matrix.

Let~$\Kunfold Y(I)$ for the given index be the \defin{unfolding} matrix of the $d$-way tensor~$\tens Y$  in the given index~$I=(\fromonetoidx kn)$, $1\leq n\leq d$, if for all its elements holds 
\begin{multline*}
    \Kunfold Y(I)[
    \overline{i_{k_1}\cdots i_{k_n}}, \,
    \overline{i_{p_1}\cdots i_{p_{d-n}}}
    ]  = 
    \tens Y[i_1,\,i_2,\,\ldots,\,i_d],
    \\
    \{\fromonetoidx p{d-n}\} =\{\fromoneto d\}\setminus I.
\end{multline*}
By a line on a group of indices, we mean a multi-index composed of the given indices, \ie the position of the corresponding sequence of indices in the list of all possible values.
We do not fix a particular sorting type of this sequence (lexicographic order can be taken) since the rearrangement
does not affect the rank of the matrix or the property of its submatrix of maximal volume.


For a non-leaf node~$\nN lj$, the node up indices~$\nid lj$ and up and down indices values~$\nvu lj$ and ~$\nvd lj$ we can construct the unfolding~$\Kunfold Y_{l,\,j}$ as
$ 
    \matr Y_{l,\,j}=
    \Kunfold Y(\niu lj).
$ 
If we consider submatrix~$\matr Y_{l,\,j}[\nvu lj,\,\nvd lj]\in\dims{r\times r}$
of this matrix based on the values~$\nvu lj$ and ~$\nvd lj$,
where $r=\nr lj$ is the corresponding rank, when our goal is to make a volume of the matrix~$\matr Y_{l,\,j}[\nvu lj,\,\nvd lj]$
as large as possible by choosing indices values~$\nvu lj$ and~$\nvd lj$.
Recall that the \defin{volume} of any (tall) matrix~$A$ is defined as
\begin{equation*}
    \vol A=\sqrt{\det A^T A},\quad A\in\dims{n\times m},
    \quad n\geq m.
\end{equation*}

\subsection{Index Values Update Algorithm}

\begin{algorithm}[tbh!]
	\caption{Indices values update algorithm.}
    \label{alg:update idx}
	\begin{algorithmic}[1]
        \REQUIRE{function~$f$ for the $d$-way tensor value calculation; indices~$i$, $i_1$ and~$i_2$ such that $\setop(i\cup i_1\cup i_2)=\{\fromoneto d$\},
        and the corresponding indices values~$v$, $v_1$ and~$v_2$; possible maximum rank increment~$\Delta r$; threshold for rank reduction~$\epsilon$; transformation~$T$.}
        \ENSURE{indices values~$V$ of the index~$I=i_1\cup i_2$.
    }
        \STATE{$r,\, r_1,\, r_2 = |v|,\, |v_1|,\,|v_2|$}
        \STATE{\comm{First we build the tall matrix:}}
        \STATE{$A\gets zeros([r_1\cdot r_2, \, r])$ \comm{Matrix with tensor values}}
        \STATE{$J\gets zeros(d)$ \comm{Integer index vector}}
        \STATE{$F\gets zeros([r_1\cdot r_2, \, r_1+r_2])$ \comm{Stores index candidates}}
        \FOR{$(j_1, j_2)$ in $\{\fromoneto r_1\}\times\{\fromoneto r_2\}$}
        \STATE{$J[i_1]\gets v_1[j_1]$}
        \STATE{$J[i_2]\gets v_2[j_2]$}
        \STATE{$F[\overline{j_1j_2},\,:\,]\gets v_1[j_1]\cup v_2[j_2]$}
        \FOR{$j$ in $\{\fromoneto r\}$}
        \STATE{$J[i]\gets v[j]$}
        \STATE{$A[\overline{j_1j_2},\,j]\gets f(J)$}
        \ENDFOR
        \ENDFOR
        \STATE{\comm{Save values of A as a cache for future use}}
        \STATE{$A\gets T(A)$ \comm{Apply point-wise transformation}}
        \STATE{\comm{Now use maxvol to select indices}}
        \STATE{$\{Q,\,R\,,P\}\gets \qrp(A)$ \comm{QR with permutations}}
        \STATE{$r_\epsilon\gets \max\{n\!\mid1\leq n\leq r,\,R[n,\,n]/R[0,\,0]\geq\epsilon\} $}
        \IF{$r_\epsilon<r$}
        \STATE{\comm{Decrease in rank occur, there is no point in raising it back again}}
        \STATE{$\Delta r\gets0$}
        \STATE{$Q\gets Q[\,:\,,\, 1{:}r_\epsilon]$}
        \ENDIF
        \STATE{$N\gets\maxvol(Q,\,\Delta r)$ \comm{$N$ is a vector of integers of length~$r_o$, $r_\epsilon\leq r_o\leq r_\epsilon+\Delta r$}}
        
        \fat{return} $F[N, \,:\,]$ \comm{List of $r_o$ vectors of length $(r_1+r_2)$ 
        }
	\end{algorithmic}
\end{algorithm}

While our method is running, we update all indices (both up and down), using the same procedure, as presented in Algorithm~\ref{alg:update idx}.
Here the function $zeros$ reserves the specified number of elements for a vector, matrix, etc.
The $\qrp$ function returns a QR decomposition with permutations (\ie, the elements on the diagonal of~$R$ do not decrease; we use the implementation from the Python package \pack{scipy}).
The operation of the algorithm can be briefly described as follows.
We construct a tall matrix, whose rows correspond to the tensor product of index values, which are conditionally called ``incoming'' and columns to ``outgoing'' ones.
Then, using the $\maxvol$ procedure, we select rows from this matrix so that the submatrix corresponding to them is of quasi-maximal volume, and the index values corresponding to these rows are returned.

\paragraph{Algorithm Input Indices.}

The ``incoming'' $i_1$, $i_2$, and ``outgoing'' $i$ input indices in our algorithm depend on the indices that are updated at each step (see \cref{fig:update idx}).
Namely, if we update the upper indices values~$\nvu lj$ for some node~$\nN lj$, then ``incoming'' indices are the upper indices of children~\childa and~\childb of this node: $i_1\gets\niu {l+1}{j_1}$, $i_2\gets\niu {l+1}{j_2}$.
The ``output'' indices are down indices of the node~$\nN lj$:
$i\gets\nid lj$ (see \cref{fig:update idx up}).
If, in turn, we update the values of down indices for the link that connects parent~$\nN lj$ and child~\childa, then for ``incoming'' indices we have: $i_1\gets\nid {l}{j}$, $i_2\gets\niu {l+1}{j_2}$, where $j_2$ is the number of another child~\childb of the node~$\nN lj$ (which differs from the original child~\childa).
The upper indices of the given link are the ``output'' ones: $i\gets\niu {l+1}{j_1}$ (see \cref{fig:update idx down}).

\paragraph{Transformation of the Tensor Values.}

The point-wise transformation is needed when we search for the minimum.
In this case, we can transform tensor values by any monotonic decreasing function~$T$.
In our experiments, we use the following adaptive (\ie its parameters dependent on the given data batch) transformation~$T$
\begin{gather*}
    T(x)=\exp\bigl(-(x-x_0)/\sigma\bigr),\\
    x_0=\mean (x),\quad
    \sigma=\std (x),\nonumber
\end{gather*}
where $\mean (x)$ and $\std (x)$ are the sample mean and sample variance of the set of numbers respectively.
When searching for the maximum value, we do a similar transformation (note that transformation can be avoided in this case; however, we apply it for greater stability of the method)
\begin{equation*}
    T(x)=\exp\bigl((x-x_0)/\sigma\bigr).
\end{equation*}

\paragraph{MaxVol Procedure.}

The $\maxvol$ procedure in our algorithm is the so-called \defin{rectangular} maximum volume search method~\cite{mikhalev2018rectangular}.
Note that it can return not only square matrices but also rectangular matrices, making a decision on the number of returned rows based on a heuristic procedure based on the possible increase in volume when adding a candidate row and the given tuning parameters.
In our numerical experiments, we allowed to expand output index set by at most $\Delta r=1$ element, so the ranks grew by at most 1 per pass.

\subsection{Traversal Procedure}

When updating indices, we walk sequentially to the neighboring (linked) node, going back only if we reach a leaf node.
As each visited node, we increment its visit counter by one, whereas at the beginning all counters were reset to zero.
When we pass through an edge, we update only one set of index values at a time: if we go from parent to child, we update down indices values; if we go from child to parent, we update upper indices values.

To decide which of the two nodes to take the next step to (in case there are two options), we count the average number of visits in each part of the tree that separates each of the two paths.
Namely, we cut the edge that was traveled last, and we cut the edges connecting the current node to the two candidate nodes.
Since by definition there are no loops in the tree and each edge is a cut edge, we get three components of connectedness.
Then we calculate the average number of visits (the sum of the number of visits on all nodes divided by the number of nodes) in each of the two connectivity components and go where the number is smaller.
If the average number of visits is close, namely, they differ by no more than a given~$\alpha$ value, then we go to the random side.
Please, see \cref{fig:path} for example of path, where each number in a list inside a node (blue circle) represents the number of steps when updates occur in this node.

\subsection{Cores Building}

After all indices are found by the search procedure described above, we can build all cores based on these indices.
First, consider a leaf node~$\nN Lj$, and let its down indices are $\nid Lj$ and the values of these indices are $\nvd Lj$ (recall, that $\{j\}\cup\nid Lj=\{\fromoneto d\}$).
Then the core, associated with this node is calculated as follows.
First, we form a matrix~$V$ of values of the BB using these indices
\begin{equation*}
\begin{split}
    \matval Vik=
    f(I_{ik})
    \;\text{ with}\quad
    I_{ik}[j]=i,
    \;\;
    \\
    I_{ik}[\nid Lj]=\nvdsame[k],
    \;\;
    \forall
    1\leq i\leq N_j,\;
    1\leq k\leq \nr Lj,
\end{split}
\end{equation*}
and then we let the core~$\G Lj$ be the transposed factor $Q$ of the QR-decomposition of this matrix~$V$
\begin{equation*}
    \G Lj = \trans Q,
    \text{ where }
    \{Q,\, R\} = \qr( V ).
\end{equation*}

For the non-leaf and non-root node~$\nN lj$ we perform a similar procedure.
Let $\nid lj$ and $\nvd lj$ be its down indices and its down indices value, respectively.
Let $\niu {l+1}{j_c}$ and $\nvusame$ be upper indices and upper indices value, respectively, for the $c$th child of this node, where $c=1,2$ (recall, that $\niu {l+1}{j_1}\cup\niu {l+1}{j_2}\cup\nid lj=\{\fromoneto d\}$).
Then we first build the matrix~$V$
\begin{multline*}
    \matval V{\overline{in}}k=
    f(I_{ink})
    \;\text{ with}
    \\
    \def\tmpo#1#2{ I_{ink}[\niu {l+1}{j_#1}]=\nvusame[#2],}
\tmpo1i\;\;\tmpo2n
    \\
    I_{ink}[\nid lj]=\nvd lj[k],
    \\
    \forall
    \def\tmpo#1#2{1\leq #2\leq \nr {l+1}{j_#1},\;}
\tmpo1i\tmpo2n
    1\leq k\leq \nr lj,
\end{multline*}
and then we let the values of the core~$\G lj$ be the ``reshaped'' values of the factor $Q$ of the QR-decomposition of $V$
\begin{equation*}
    \G lj[i,\,k,\,n] =
    \matval Q{\overline{in}}k,
    \text{ where }
    \{Q,\, R\} = \qr( V ).
\end{equation*}

Finally, for the root node~$\nN 11$, we let the values of the assigned core be the values of the given BB in the corresponding points.
Namely, let $\niu{2}{j_c}$ and $\nvu{2}{j_c}$ be upper indices and upper indices value for the $c$th child of the root node, $c=1,2$.
Then for all $
\def\tmpo#1#2{1\leq #2\leq \nr {2}{j_#1}}
\tmpo1i,\;\tmpo2n$ we have
\def\tmpo#1#2{ I_{in}[\niu {2}{j_#1}]=\nvu {2}{j_#1}[#2]}
\begin{multline*}
    \G 11[i,\,1,\,n] = 
    \matval Qin,\;\;
    \matval Qin=
    f(I_{in})
    \\
    \text{ with}\quad
\tmpo1i,\;
\tmpo2n.
\end{multline*}
Note, that due to this procedure, the obtained cores are orthogonalized and, therefore, their maximum modulo values are moderated.

\section{Related Work}
    \label{sec:related}
    In many practical situations, the problem-specific target function is not differentiable, too complex, or its gradients are not helpful due to the non-convex nature of the problem, and it has to be treated as a black box (BB).
In this case, two important problems naturally arise: approximation~\cite{bhosekar2018advances} and optimization~\cite{alarie2021two}.

The approximation carried out in the offline phase allows us to build a surrogate (simplified) model of the BB, which can then be used in the online phase to quickly calculate its values and various characteristics.
In the multidimensional case, it becomes difficult to construct a surrogate model, and low-rank tensor approximations are often the most effective.
Several recent works~\cite{kapushev2020tensor, ahmadiasl2021cross, chertkov2023black} proposed various new algorithms based on the TT-decomposition for approximating high-dimensional functions.
If we have access to the BB and can perform dynamic queries, then the powerful TT-cross method~\cite{oseledets2010ttcross} is often used, and if only a training dataset is available, then the TT-ALS method~\cite{holtz2012alternating} is preferred.
In this work, we consider the case of adaptive queries to the BB, so we select the TT-cross method as the main baseline for the approximation problem.

Gradients are not available for the BB, so only gradient-free methods can be used for the optimization problem. Particle Swarm Optimization (PSO)~\cite{kennedy1995particle} and Simultaneous Perturbation Stochastic Approximation (SPSA)~\cite{maryak2001global} are rather useful methods in this case.
There is also a large variety of other heuristic methods for finding the global extremum.
Recently, the TT-decomposition has been actively used for black-box optimization, since it turns out to be more effective than standard approaches in the multidimensional case.
An iterative method TTOpt based on the maximum volume approach is proposed in the work~\cite{sozykin2022ttopt}.
The authors applied this approach to the problem of optimizing the weights of neural networks in the framework of reinforcement learning problems in~\cite{sozykin2022ttopt} and to the QUBO problem in~\cite{nikitin2022quantum}.
A similar optimization approach was also considered in~\cite{selvanayagam2022global} and~\cite{shetty2022tensor}.
One more promising algorithm, Optima-TT, which is based on the probabilistic sampling from the TT-tensor, was proposed in recent work~\cite{chertkov2023tensor}.
We also note the work~\cite{soley2021iterative}, where an optimization method based on the iterative power algorithm in terms of the quantized version of the TT-decomposition is proposed.
As a result, we consider classical PSO and SPSA methods as well as the TTOpt method as baselines for the optimization problem.
\section{Numerical Experiments}
    \label{sec:experiments}
    \begin{table*}[th!b]
\caption{
    Benchmark functions for performance analysis of the proposed method.}
\label{tbl:benchmarks_func}
\begin{center}
\begin{small}
\begin{sc}
\renewcommand{\arraystretch}{2.0}
\begin{tabular}{|p{2.5cm}|p{2.6cm}|p{11.0cm}|}\hline
\toprule
Function & Bounds & Analytical formula
\\ 

\midrule


Alpine
    & [$-10$, $10$]
    &
    $
    \sum_{i=1}^{d}
        | x_i \sin{x_i} + 0.1 x_i |
    $
    \\ \hline

Chung
    & [$-10$, $10$]
    &
    $
    \left(
        \sum_{i=1}^{d} x_i^2
    \right)^2
    $
    \\ \hline

Dixon
    & [$-10$, $10$]
    &
    $
    \ff(\vx)
    =
    (x_1-1)^2 +
    \sum_{i=2}^{d}
        i \cdot \left(
            2 x_i^2 - x_{i-1} \right)^2
    $
    \\ \hline

Griewank
    & [$-100$, $100$]
    &
    $
    \sum_{i=1}^d \frac{x_i^2}{4000} -
    \prod_{i=1}^d \cos{\left(\frac{x_i}{\sqrt{i}}\right)} + 1
    $
    \\ \hline

Pathological
    & [$-100$, $100$]
    &
    $
    \sum_{i=1}^{d-1} \left( 0.5 + \frac
        {
            \sin^2 \sqrt{
                100 x_i^2 + x_{i+1}^2
            } - 0.5
        }
        {
            1 + 0.001 \left(
                x_i^2 -
                2 x_i x_{i+1} +
                x_{i+1}^2
            \right)^2
        } \right)
    $
    \\ \hline

Pinter
    & [$-10$, $10$]
    &
    $
    \sum_{i=1}^{d} \left(
        i x_i^2 +
        20 i \sin^2{A_i} +
        i \log_{10}{(1 + i B_i^2)}
    \right)
    $,
    \newline
    where
    $
    A_i =
        x_{i-1} \sin{x_i} +
        \sin{x_{i+1}}
    $,
    $
    B_i = x_{i-1}^2 - 2 x_i +
        3 x_{i+1} - \cos{x_i} + 1)
    $
    \newline
    with $x_0 = x_d$ and $x_{d+1} = x_1$
    \\ \hline

Qing
    & [$0$, $500$]
    &
    $
    \ff(\vx)
    = \sum_{i=1}^d
        \left( x_i^2 - i \right)^2
    $
    \\ \hline

Rastrigin
    & [$-5.12$, $5.12$]
    &
    $
    A \cdot d + \sum_{i=1}^{d} \left(
        x_i^2 -
        A \cdot \cos{(2 \pi x_i)}
    \right),
    $
    \newline
    where $A = 10$
    \\ \hline

Schaffer
    & [$-100$, $100$]
    &
    $
    \sum_{i=1}^{d-1} \left(
        0.5 +
        \frac{
            \sin^2{\left( \sqrt{x_i^2 + x_{i+1}^2} \right)} - 0.5
        }{
            \left(
                1 +
                0.001 \cdot (
                    x_i^2 + x_{i+1}^2)
            \right)^2
        }
    \right)
    $
    \\ \hline

Schwefel
    & [$0$, $500$]
    &
    $
    -\frac{1}{d} \sum_{i=1}^d x_i \cdot \sin{(\sqrt{|x_i|})}
    $
    \\ \hline

Sphere
    & [$-5.12$, $5.12$]
    &
    $
    \sum_{i=1}^{d} x_i^2
    $
    \\ \hline

Squares
    & [$-10$, $10$]
    &
    $
    \sum_{i=1}^{d} i x_i^2
    $
    \\ \hline

Trigonometric
    & [$0$, $\pi$]
    &
    $
    \sum_{i=1}^{d} \left(
        d -
        \sum_{j=1}^{d} \cos{x_j} +
        i (1 - \cos{x_i} - \sin{x_i})
    \right)^2
    $
    \\ \hline

Wavy
    & [$-\pi$, $\pi$]
    &
    $
    1 - \frac{1}{d} \sum_{i=1}^{d}
        \cos{(k x_i)}
        \cdot
        e^{-\frac{x_i^2}{2}}
    $,
    \newline
    where $k = 10$
    \\ 

\bottomrule

\end{tabular}
\end{sc}
\end{small}
\end{center}
\vskip -0.1in
\end{table*}

\begin{table}[t!]
\caption{
    Approximation relative error for the HTBB and TT-cross applied to all considered $d = 256$-dimensional benchmarks.
    The reported values are averaged over $10$ independent runs.}
\label{tbl:results_func_appr}
\begin{center}
\begin{small}
\begin{sc}
\begin{tabular}{|p{3.0cm}|p{2.0cm}|p{2.0cm}|} 
\toprule
Benchmark &
HTBB &
TT-cross
\\ 

\midrule

Alpine
    & \fat{2.83e-15}
    & 1.73e-02
\\ 

Chung
    & \fat{7.87e-03}
    & 2.86e-02
\\ 

Dixon
    & \fat{5.65e-03}
    & 1.00e-01
\\ 

Griewank
    & \fat{2.83e-15}
    & 1.43e-02
\\ 

Pathological
    & \fat{3.92e-02}
    & 1.08e-01
\\ 

Pinter
    & \fat{1.23e-02}
    & 1.47e-02
\\ 

Qing
    & \fat{3.67e-02}
    & 4.87e-02
\\ 

Rastrigin
    & \fat{1.01e-14}
    & 1.47e-02
\\ 

Schaffer
    & \fat{1.87e-02}
    & 1.88e-02
\\ 

Schwefel
    & \fat{3.39e-14}
    & 6.31e-01
\\ 

Sphere
    & \fat{1.20e-14}
    & 1.44e-02
\\ 

Squares
    & \fat{1.07e-14}
    & 1.77e-02
\\ 

Trigonometric
    & \fat{2.76e-02}
    & 4.82e-02
\\ 

Wavy
    & \fat{8.56e-05}
    & 2.46e-03
\\ 

\bottomrule

\end{tabular}
\end{sc}
\end{small}
\end{center}
\vskip -0.1in
\end{table}

\begin{table}[t!]
\caption{
    Approximation relative error for the HTBB applied to all considered $512$ and $1024$-dimensional benchmarks.
    The reported values are averaged over $5$ independent runs.}
\label{tbl:results_func_appr_big}
\begin{center}
\begin{small}
\begin{sc}
\begin{tabular}{|p{3.0cm}|p{2.0cm}|p{2.0cm}|} 

\toprule

Benchmark  &
$d =  512$ &
$d = 1024$
\\ 

\midrule

Alpine
    & 4.92e-15
    & 3.81e-04
\\

Chung
    & 7.86e-03
    & 7.64e-03
\\

Dixon
    & 3.75e-03
    & 2.83e-03
\\

Griewank
    & 1.37e-14
    & 3.16e-14
\\

Pathological
    & 3.80e-02
    & 3.76e-02
\\

Pinter
    & 8.80e-03
    & 8.38e-03
\\

Qing
    & 1.85e-02
    & 1.60e-02
\\

Rastrigin
    & 1.63e-14
    & 1.02e-04
\\ 

Schaffer
    & 1.94e-02
    & 1.52e-02
\\ 

Schwefel
    & 2.59e-13
    & 1.23e-13
\\ 

Sphere
    & 1.16e-14
    & 4.58e-14
\\ 

Squares
    & 1.08e-14
    & 2.38e-14
\\ 

Trigonometric
    & 2.74e-02
    & 2.38e-02
\\ 

Wavy
    & 1.18e-04
    & 3.38e-04
\\ 

\bottomrule

\end{tabular}
\end{sc}
\end{small}
\end{center}
\vskip -0.1in
\end{table}

\begin{table*}[t!]
\caption{
    Minimization results for the HTBB, TTOpt, One+One, SPSA, and PSO applied to all considered $256$-dimensional benchmarks.
    The reported values are averaged over $10$ independent runs.}    
\label{tbl:results_func_opti}
\begin{center}
\begin{small}
\begin{sc}
\begin{tabular}{|p{3.1cm}|p{2.3cm}|p{2.3cm}|p{2.3cm}|p{2.3cm}|p{2.3cm}|} 

\toprule

Benchmark
& HTBB
& TTOpt
& One+One
& SPSA
& PSO
\\ 

\midrule

Alpine
    & \fat{6.75e+01}
    & 4.48e+02
    & 3.66e+02
    & 3.99e+02
    & 4.76e+02
\\ 

Chung
    & \fat{1.45e+06}
    & 7.74e+07
    & 1.48e+06
    & 1.54e+06
    & 6.98e+07
\\ 

Dixon
    & \fat{1.89e+06}
    & 1.99e+08
    & 2.33e+06
    & 3.20e+06
    & 2.68e+08
\\ 

Griewank
    & \fat{3.11e+01}
    & 2.21e+02
    & 3.19e+01
    & 3.12e+01
    & 2.09e+02
\\ 

Pathological
    & \fat{6.97e+01}
    & 1.02e+02
    & 1.14e+02
    & 9.32e+01
    & 1.06e+02
\\ 

Pinter
    & \fat{5.17e+05}
    & 1.19e+06
    & 5.67e+05
    & 5.94e+05
    & 1.51e+06
\\ 

Qing
    & \fat{4.99e+06}
    & 2.98e+12
    & 8.47e+10
    & 1.26e+12
    & 1.76e+12
\\ 

Rastrigin
    & \fat{9.19e+02}
    & 3.61e+03
    & 1.09e+03
    & 9.31e+02
    & 3.72e+03
\\ 

Schaffer
    & \fat{9.87e+01}
    & 1.15e+02
    & 1.06e+02
    & 1.02e+02
    & 1.20e+02
\\ 

Schwefel
    & \fat{-3.85e+02}
    & -1.77e+02
    & -1.92e+02
    & -1.89e+02
    & -1.38e+02
\\ 

Sphere
    & \fat{3.16e+02}
    & 2.30e+03
    & 3.24e+02
    & 3.17e+02
    & 2.19e+03
\\ 

Squares
    & \fat{1.55e+05}
    & 7.37e+05
    & 1.57e+05
    & \fat{1.55e+05}
    & 1.02e+06
\\ 

Trigonometric
    & \fat{8.72e+04}
    & 9.30e+06
    & 2.62e+05
    & 1.77e+07
    & 1.01e+07
\\ 

Wavy
    & \fat{3.19e-01}
    & 6.21e-01
    & 3.64e-01
    & 3.22e-01
    & 6.36e-01
\\ 

\bottomrule

\end{tabular}
\end{sc}
\end{small}
\end{center}
\vskip -0.1in
\end{table*}

To demonstrate the effectiveness of the proposed HTBB approach, we select $14$ popular $256$-dimensional benchmarks~\cite{jamil2013literature,vanaret2020certified,dieterich2012empirical}, which correspond to analytical functions with complex landscape and are described in detail in \cref{tbl:benchmarks_func}.
For each benchmark, we fix the input dimension at $256$ and consider the approximation and optimization problem in the black-box settings for the tensor that arises when the corresponding function is discretized on a Chebyshev grid with $8$ nodes in each dimension.
In all cases, we limited the budget (the number of requests to the BB) to $10^4$, and the HT-rank was taken to be $2$.

\subsection{Multidimensional approximation}

\begin{figure}[t!]
    \vskip 0.2in
    \centering
    \hfill
    \begin{subfigure}{0.49\linewidth}
        \includegraphics[width=0.99\linewidth]{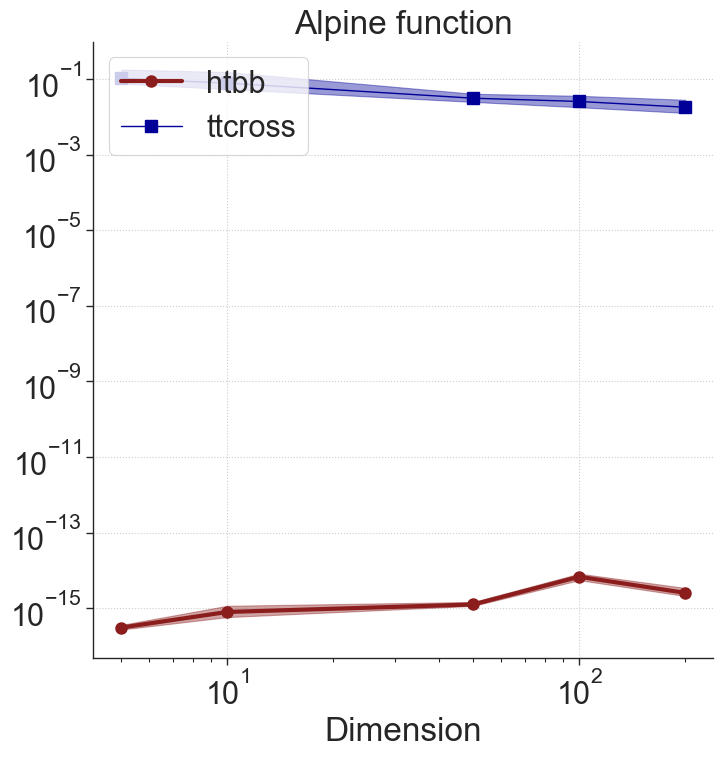}
    \end{subfigure}
    \begin{subfigure}{0.49\linewidth}
        \includegraphics[width=0.99\linewidth]{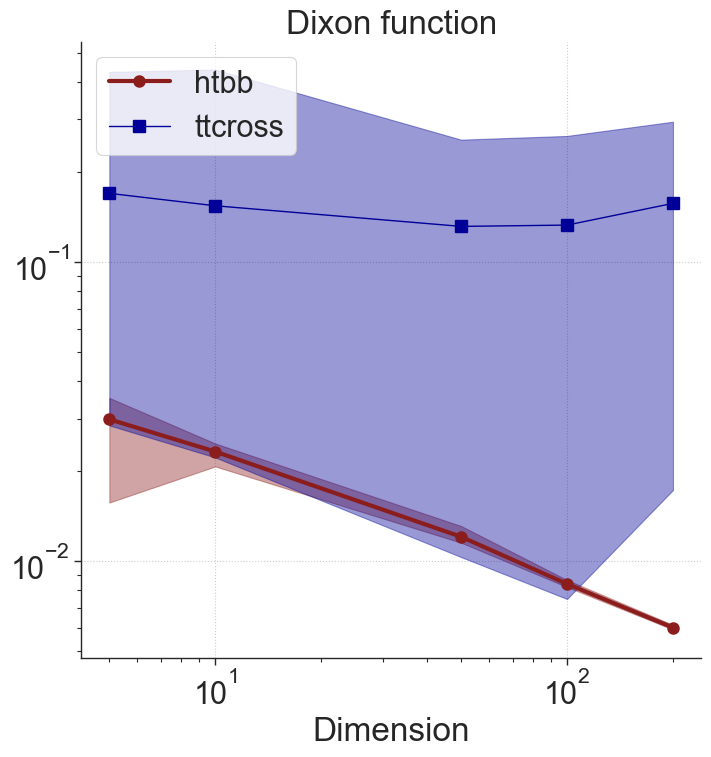}
    \end{subfigure}
    \caption{
        Approximation results for Alpine and Dixon functions for cases of dimensions $5$, $10$, $50$, $100$, and $200$.
        For both methods, we plot the relative error of the solution averaged over $10$ runs with a solid line and fill in the area between the worst and best result with the same color.
    }
    \label{fig:results_appr}
    \vskip -0.2in
\end{figure}

For each $256$-dimensional benchmark we perform the approximation with the proposed HTBB method and compare it with the TT-cross method,\footnote{
    We used the implementation of the TT-cross method from \url{https://github.com/AndreiChertkov/teneva}.
} constrained by the same budget ($10^4$ requests to BB).
The relative $L2$ errors on test sets of $10^4$ random points which were generated for each benchmark are reported in \cref{tbl:results_func_appr} (the computations were repeated $10$ times for both methods and the averaged results are presented).
Also in \cref{fig:results_appr} we provide a graphical comparison of the results for two benchmarks for the case of different values of the problem dimension ($5$, $10$, $50$, $100$, $200$).
As follows from the presented results, for all problems our method turns out to be more accurate than the baseline, and in some cases its accuracy turns out to be many orders of magnitude higher.
For the case of higher dimensions for the considered problem classes, running the TT-cross method leads to failures in software implementation due to instability, while our approach remains stable and gives high accuracy, as follows from values reported in \cref{tbl:results_func_appr_big} for dimensions $512$ and $1024$.

\subsection{Multidimensional optimization}

\begin{figure}[t!]
    \vskip 0.2in
    \centering
    \begin{subfigure}{0.49\linewidth}
        \includegraphics[width=0.99\linewidth]{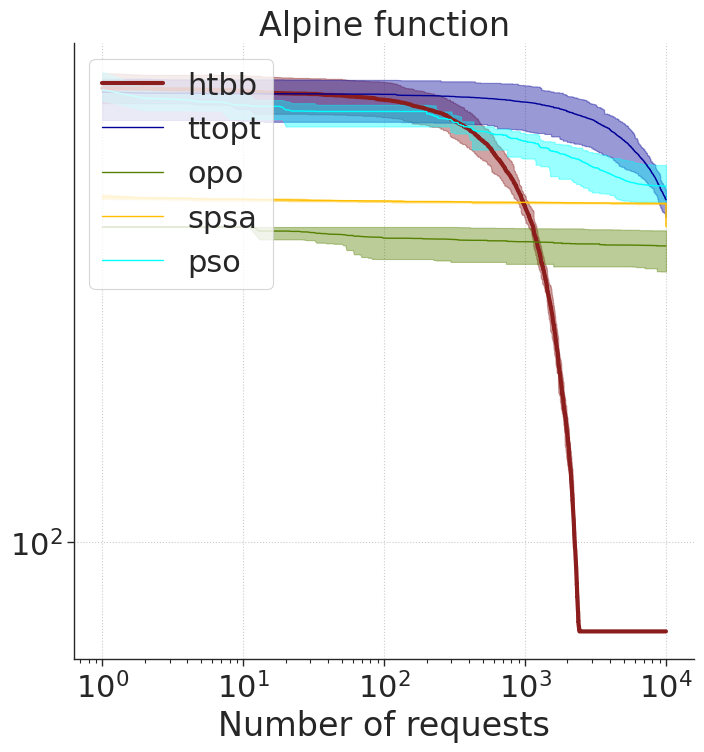}
    \end{subfigure}
    \begin{subfigure}{0.49\linewidth}
        \includegraphics[width=0.99\linewidth]{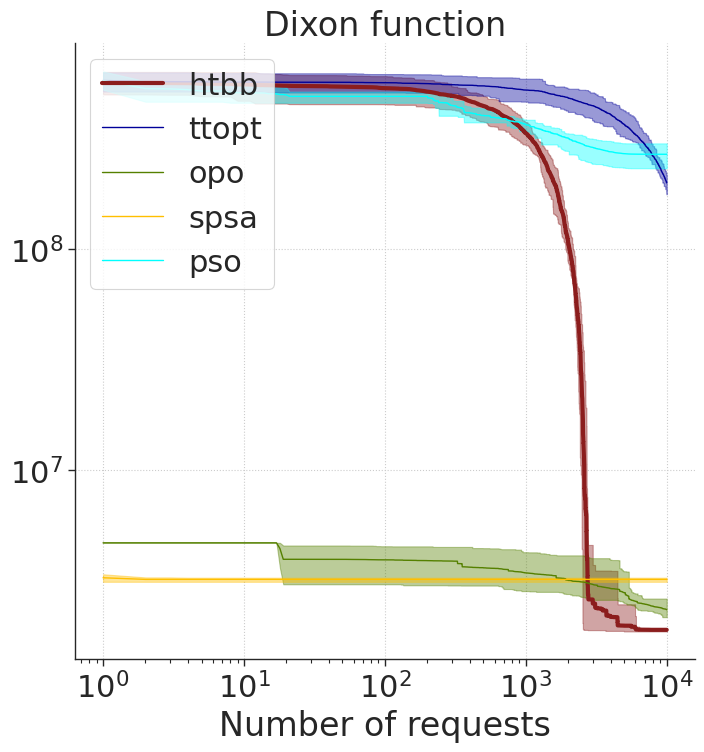}
    \end{subfigure}
    \caption{
        Minimization results for Alpine and Dixon functions.
        For each of the optimizers, we plot the value of the solution averaged over $10$ runs with a solid line and fill in the area between the worst and best result with the same color.
    }
    \label{fig:results_opti}
    \vskip -0.2in
\end{figure}

For each $256$-dimensional benchmark we perform the optimization (namely the search for a global minimum) with the proposed HTBB method.
We consider as baselines the tensor-based optimization method TTOpt\footnote{
    We used the implementation of the TTOpt method from \url{https://github.com/AndreiChertkov/ttopt}.
} and three popular gradient-free optimization algorithms from the nevergrad framework~\cite{bennet2021nevergrad}:\footnote{
    See \url{https://github.com/facebookresearch/nevergrad}.
} One+One, SPSA, and PSO.
The limit on the number of requests to the objective function was fixed at the value  $10^4$.
The calculations were repeated $10$ times for all methods and the averaged results are presented in \cref{tbl:results_func_opti}.
Also in \cref{fig:results_opti} we show the convergence plots for two benchmarks.
As follows from the reported values, HTBB, in contrast to alternative approaches, gives a consistently top result for all model problems.

\section{Conclusions}
    \label{sec:conclusions}
    In this work, we presented a new method HTBB for simultaneously solving the problem of multidimensional approximation and gradient-free optimization for functions given in the form of a black box.
Our approach is based on the low-rank hierarchical Tucker decomposition, which makes it especially effective in the multidimensional case.
The key features of the presented work are a)~using the MaxVol algorithm which allows efficiently finding the required indices 
and b)~using the sequential traversal of cores,
allowing to move to one of the neighboring nodes and making it more efficient to find indexes that need updating.

The HTBB method can be applied to a wide class of practically significant problems,
including optimal control and various machine learning applications.
As future work, we point out the possibility of a rather simple extension on the HT-structure of the algorithms that now exist for the TT-decomposition: rounding, orthogonalization, search for the maximum element by the top-k-like methods, etc.

\bibliography{biblio}
\bibliographystyle{icml2024}


\end{document}